\documentclass[a4paper,final]{article}
\usepackage[dvips]{graphicx}
\usepackage{amsmath} 
\usepackage{amssymb} 

\makeatletter
\def\@maketitle{
  \vbox to 1.5in{
   \hsize\textwidth
    \linewidth\hsize
    \vspace*{1.5cm}
    \centering
    {\bfseries\huge \@title \par}
    \vskip 2em
    {\large \begin{tabular}[t]{c}\@author \end{tabular}\par}
    \vfill}    \vspace*{1.0cm}
}
\renewcommand\section{\@startsection {section}{1}{\z@}
     {.7\baselineskip plus\baselineskip}{.5\baselineskip}
                                   {\normalfont\Large\bfseries}}
\renewcommand\section{\@startsection {section}{1}{\z@}
      {.5\baselineskip\@plus.7\baselineskip}{.3\baselineskip}
                                   {\normalfont\Large\bfseries}}
\renewcommand\subsection{\@startsection{subsection}{2}{\z@}
       {.5\baselineskip\@plus.7\baselineskip}{.3\baselineskip}
                                   {\normalfont\large\bfseries}}
\renewcommand\subsubsection{\@startsection{subsubsection}{3}{\z@}
      {.5\baselineskip\@plus.7\baselineskip}{.3\baselineskip}
                                     {\normalfont\normalsize\bfseries}}
\makeatother

\renewenvironment{abstract}
  {\normalfont
    \list{}{\labelwidth0pt
      \leftmargin0pt \rightmargin\leftmargin
      \listparindent\parindent \itemindent0pt
      \parsep0pt
      
    }
    \item[\hskip\labelsep\bfseries\abstractname\enspace --] \itshape
}{
  \endlist}

\newcommand{\keywordsname}{Keywords}
\newenvironment{keywords}
  {\normalfont
    \list{}{\labelwidth0pt
      \leftmargin0pt \rightmargin\leftmargin
      \listparindent\parindent \itemindent0pt
      \parsep0pt
      }
    \item[\hskip\labelsep\bfseries\keywordsname:]}{\endlist}

\begin{document}

\pagestyle{myheadings}
\markboth{}{}

\title{Fusion of qualitative beliefs using DSmT}

\author{
Florentin Smarandache\\
Department of Mathematics\\
University of New Mexico\\
Gallup, NM 87301, U.S.A.\\
smarand@unm.edu\\
\and Jean Dezert\\
ONERA\\
29 Av. de la  Division Leclerc \\
92320 Ch\^{a}tillon, France.\\
Jean.Dezert@onera.fr
}

\date{}

\maketitle

\begin{abstract}
This paper introduces the notion of qualitative belief assignment to model beliefs of human experts expressed in natural language (with linguistic labels). We show how qualitative beliefs can be efficiently combined using an extension of Dezert-Smarandache Theory (DSmT) of plausible and paradoxical quantitative reasoning to qualitative reasoning. We propose a new arithmetic on linguistic labels which allows a direct extension of classical DSm fusion rule or DSm Hybrid rules. An approximate qualitative PCR5 rule is also proposed jointly with a Qualitative Average Operator. We also show how crisp or interval mappings can be used to deal indirectly with linguistic labels. A very simple example is provided to illustrate our qualitative fusion rules.
\end{abstract}


\section{Introduction}

Since fifteen years qualitative methods for reasoning under uncertainty developed in Artificial Intelligence are attracting more and more people of Information Fusion community, specially those working in the development of modern multi-source\footnote{Where both computers, sensors and human experts are involved in the loop.} systems for defense.  Their aim is to propose solutions for processing and combining qualitative information to take into account efficiently information provided by human sources (or "semi-intelligent" expert systems) and usually expressed in natural language rather than direct quantitative information. George Polya was one of the first mathematicians to attempt a formal characterization of qualitative human reasoning in 1954 \cite{Polya_1954}, then followed by Lofti Zadeh's works \cite{Zadeh_1975}-\cite{Zadeh_2002}. The interest of qualitative reasoning methods is to help in decision-making for situations in which the precise numerical methods are not appropriate (whenever the information/input are not directly expressed in numbers). Several formalisms for qualitative reasoning have been proposed as extensions on the frames of probability, possibility and/or evidence theories \cite{Bobrow_1984,Dubois_1989,Darwiche_1992,Wong_1993,Lamata_1994,Zadeh_1996,Zadeh_2002,Yager_2004b}. The limitations of numerical techniques are discussed in \cite{Parsons_1998}. We browse here few main approaches. A detailed presentation of theses techniques can be found in \cite{Parsons_2001}. In \cite{Wellman_1994}, Wellman proposes a general characterization of  "qualitative probability" to relax precision in representation and reasoning within the probabilistic framework. His basic idea was to develop Qualitative Probabilistic Networks (QPN) based on a Qualitative Probability Language (QPL) defined by a set of numerical underlying probability distributions. The major interest of QPL is to specify the partial rankings among degrees of belief rather than assessing their magnitudes on a cardinal scale. Such method cannot be considered as truly qualitative, since it belongs to the family of imprecise probability \cite{Walley_1991} and probability bounds analysis (PBA) methods \cite{Ferson_1998}. Some advances have been done by Darwiche in \cite{Darwiche_1992} for a symbolic generalization of Probability Theory; more precisely, Darwiche proposes a support (symbolic and/or numeric) structure which contains all information able to represent and conditionalize the state of belief. Darwiche shows that Probability Theory fits within his new support structure framework as several other theories, but Demspter-Shafer Theory doesn't fit in. Based on Demspter-Shafer Theory \cite{Shafer_1976} (DST), Wong and Lingras \cite{Wong_1994} propose a method for generating a (numerical) basic belief functions from preference relations between each pair of propositions be specified qualitatively. The algorithm proposed doesn't provide however a unique solution and doesn't check the consistency of qualitative preference relations. Bryson and al. \cite{Bryson_1998,Ngwenyama_1998} propose a procedure called Qualitative Discriminant Procedure (QDP) that involves qualitative scoring, imprecise pairwise comparisons between pairs of propositions and an optimization algorithm to generate consistent imprecise quantitative belief function to combine. Very recently, Ben Yaglane in \cite{BenYaghlane_2005} has reformulated the problem of generation of quantitative (consistent) belief functions from qualitative preference relations as a more general optimization problem under additional non linear constraints in order to minimize different uncertainty measures (Bezdek's entropy, Dubois \& Prade non-specificity, etc). In \cite{Parsons_1993,Parsons_1994}, Parsons proposes a qualitative Dempster-Shafer Theory, called Qualitative Evidence Theory (QET),  by using techniques from qualitative reasoning \cite{Bobrow_1984}. Parsons' idea is to use {\it{qualitative belief assignments}} (qba), denoted here $qm(.)$ assumed to be only $0$ or $+$, where $+$ means some unknown value between 0 and 1. Parsons proposes, using operation tables, a very simple arithmetic for qualitative addition $+$ and multiplication $\times$ operators. The combination of two (or more) qba's then actually follows the classical conjunctive consensus operator based on his qualitative multiplication table. Because of impossibility of qualitative normalization, Parsons uses the un-normalized version of Dempster's rule by committing a {\it{qualitative mass}} to the empty set following the open-world approach of Smets \cite{Smets_1988}. This approach cannot deal however with truly closed-world problems because there is no issue to transfer the conflicting qualitative mass or to normalize the qualitative belief assignments in the spirit of DST. An improved version of QET has been proposed \cite{Parsons_1993} for using refined linguistic quantifiers as suggested by Dubois \& Prade in \cite{Dubois_1992}. The fusion of refined qualitative belief masses follows the un-normalized Dempster's rule based on an underlying numerical interval arithmetic associated with linguistic quantifiers. Actually, this refined QTE fits directly within DSmT framework since it corresponds to imprecise (quantitative) DSmC fusion rule \cite{DSmTBook_2004a,Dezert_2006}.  From 1995, Parsons switched back to qualitative probabilistic reasoning \cite{Parsons_1998b} and started to develop Qualitative Probabilistic Reasoner (QPR). Recently, the author discussed about the flaw discovered in QPR and gave some issues with new open questions \cite{Parsons_2004}. In Zadeh's paradigm of computing with words (CW) \cite{Zadeh_1996}-\cite{Zadeh_2002} the combination of qualitative/vague information expressed in natural language is done essentially in three steps: 1) a translation of qualitative information into fuzzy membership functions, 2) a fuzzy combination of fuzzy membership functions; 3) a retranslation of fuzzy (quantitative) result into natural language. All these steps cannot be uniquely accomplished since they depend on the fuzzy operators chosen. A possible issue for the third step is proposed in \cite{Yager_2004b}. In this paper we propose a simple arithmetic of linguistic labels which allows a direct extension of classical (quantitative) combination rules proposed in the DSmT framework into their qualitative counterpart.

\section{Qualitative Operators}

Computing with words (CW) and qualitative information is more vague, less precise than computing with numbers, but it offers the advantage of robustness if done correctly since :\\

{\it{"It would be a great mistake to suppose that vague knowledge must be false. On the contrary, a vague belief has a much better chance of being true than a precise one, because there are more possible facts that would verify it."}} - Bertrand Russell \cite{Russell_1923}.\\

We propose in this section a general arithmetic for computing with words (i.e. with linguistic labels).\\

So let's consider a finite frame $\Theta=\{\theta_1,\ldots,\theta_n\}$ of $n$ (exhaustive) elements $\theta_i$, $i=1,2,\ldots,n$, with an associated model $\mathcal{M}(\Theta)$ on $\Theta$ (either Shafer's model $\mathcal{M}^0(\Theta)$, free-DSm model $\mathcal{M}^f(\Theta)$, or more general any Hybrid-DSm model \cite{DSmTBook_2004a}). A model $\mathcal{M}(\Theta)$ is defined by the set of integrity constraints on elements of $\Theta$ (if any); Shafer's model $\mathcal{M}^0(\Theta)$ assumes all elements of $\Theta$ truly exclusive, while free-DSm model $\mathcal{M}^f(\Theta)$ assumes no exclusivity constraints between elements of the frame $\Theta$. \\

Let's define a finite set of linguistic labels\index{linguistic labels} $\tilde{L}=\{L_1,L_2,\ldots,L_m\}$ where $m\geq 2$ is an integer. $\tilde{L}$ is endowed with a total order relationship $\prec$, so that $L_1\prec L_2\prec \ldots\prec L_m$. To work on a close linguistic set under linguistic addition and multiplication operators, we extends $\tilde{L}$ with two extreme values $L_{0}$ and $L_{m+1}$ where $L_{0}$ corresponds to the minimal qualitative value and $L_{m+1}$ corresponds to the maximal qualitative value, in such a way that
$$L_0\prec L_1\prec L_2\prec \ldots\prec L_m\prec L_{m+1}$$

\noindent
where $\prec$ means inferior to, or less (in quality) than, or smaller (in quality) than, etc. hence a relation of order from a qualitative point of view. But if we make a correspondence between qualitative
labels and quantitative values on the scale $[0, 1]$, then $L_{\min}=L_0$ would correspond to the numerical value 0, while $L_{\max}=L_{m+1}$ would correspond to the
numerical value 1, and each $L_i$ would belong to $[0,1]$, i. e.

$$L_{\min}=L_0 < L_1 < L_2 < \ldots <L_m < L_{m+1}=L_{\max}$$

\noindent
From now on, we work on extended ordered set $L$ of qualitative values
$$L=\{L_0,\tilde{L},L_{m+1}\}=\{L_0,L_1,L_2,\ldots,L_m,L_{m+1}\}$$

\noindent
The qualitative addition and multiplication operators are respectively defined in the following way:

\begin{itemize}
\item Addition :
\begin{equation}
L_i + L_j=
\begin{cases}
L_{i+j}, & \text{if}\ i+j < m+1,\\
L_{m+1}, & \text{if}\ i+j \geq m+1.
\end{cases}
\label{qadd}
\end{equation}
\item Multiplication :
\begin{equation}
L_i \times L_j=L_{\min\{i,j\}}
\label{qmult}
\end{equation}
\end{itemize}

\noindent
These two operators are well-defined, commutative, associative, and unitary\index{unitary operator}. Addition of labels is a unitary operation\index{unitary operator} since $L_0 =
L_{\min}$ is the unitary element, i.e. $L_i + L_0 = L_0 + L_i = L_{i+0} = L_i$ for all $0\leq  i \leq
m+1$.
Multiplication of labels is also a unitary operation\index{unitary operator} since $L_{m+1} = L_{\max}$ is the unitary element, i.e.
$L_i \times L_{m+1} = L_{m+1} \times L_i = L_{\min\{i, m+1\}} = L_i$ for $0\leq  i \leq m+1$. $L_0$ is the unit element for addition, while $L_{m+1}$ is the unit element for multiplication. $L$ is closed under $+$ and $\times$. The mathematical structure formed by $(L, +, \times)$ is a commutative bisemigroup\index{bisemigroup} with different unitary elements for each operation. We recall that a bisemigroup\index{bisemigroup} is a set $S$ endowed with two associative binary operations such that $S$ is closed under both operations.\\

If $L$ is not an exhaustive set of qualitative labels, then other labels may exist in between the initial ones, so we can work with labels and numbers - since a
refinement of $L$ is possible. When mapping from $L$ to crisp numbers or intervals, $L_0 = 0$ and $L_{m+1}=1$, while $0<L_i<1$, for
all $i$, as crisp numbers, or $L_i$ included in $[0,1]$ as intervals/subsets.\\

For example, $L_1$, $L_2$, $L_3$ and $L_4$ may represent the following qualitative values: $L_1\triangleq \text{very poor}$, $L_2\triangleq \text{poor}$, $L_3\triangleq \text{good}$ and $L_4\triangleq \text{very good}$ where $\triangleq$ symbol means "by definition".\\

We think it is better to define the multiplication $\times$ of $L_i\times L_j$ by $L_{\min\{i,j\}}$ because multiplying two numbers $a$ and $b$ in $[0,1]$ one gets a result which is less than each of them, the product is not bigger than both of them as Bolanos et al. did in \cite{Bolanos_1993} by approximating $L_i\times L_j= L_{i+j} > \max\{L_i, L_j\}$.  While for the addition it is the opposite: adding two numbers in the interval $[0,1]$ the sum should be bigger than both of them, not smaller as in \cite{Bolanos_1993} case where $L_i+L_j = \min\{L_i, L_j\}< \max\{L_i,
L_j\}$.

\section{Qualitative Belief Assignment}
We define a qualitative belief assignment\index{qualitative belief assignment} (qba), and we call it {\it{qualitative belief mass}} or {\it{q-mass}} for short, a mapping function $qm(.): G^\Theta \mapsto L$ where $G^\Theta$ corresponds the space of propositions generated with $\cap$ and $\cup$ operators and elements of $\Theta$ taking into account the integrity constraints of the model. For example if Shafer's model is chosen for $\Theta$, then $G^\Theta$ is nothing but the classical power set $2^\Theta$ \cite{Shafer_1976}, whereas if free DSm model is adopted $G^\Theta$ will correspond to Dedekind's lattice (hyper-power set) $D^\Theta$ \cite{DSmTBook_2004a}. Note that in this qualitative framework, there is no way to define normalized $qm(.)$, but qualitative quasi-normalization\index{quasi-normalization} is still possible as seen further. Using the qualitative operations defined previously we can easily extend the combination rules from quantitative
to qualitative. In the sequel we will consider $s\geq 2$ qualitative belief assignments\index{qualitative belief assignment} $qm_1(.),\ldots, qm_s(.)$ defined over the same space $G^\Theta$ and provided by $s$ independent sources $S_1,\ldots,S_s$ of evidence. \\

 \noindent{\bf{Important note}}: The addition and multiplication operators used in all qualitative fusion formulas in next sections correspond to {\it{qualitative addition}} and {\it{qualitative multiplication}} operators defined in \eqref{qadd} and \eqref{qmult} and must not be confused with classical addition and multiplication operators for numbers.
 
\section{Qualitative Conjunctive Rule}

The qualitative Conjunctive Rule (qCR)\index{qualitative Conjunctive Rule (qCR)} of $s\geq 2$ sources is defined similarly to the quantitative conjunctive consensus rule, i.e.

\begin{equation}
qm_{qCR}(X)=\sum_{\substack{X_1,\ldots,X_s\in G^\Theta\\ X_1\cap \ldots \cap X_s=X}} \prod_{i=1}^{s} qm_i(X_i)
\label{qCR}
\end{equation}

\noindent
The total qualitative conflicting mass is given by $$K_{1\ldots s}=\sum_{\substack{X_1,\ldots,X_s\in G^\Theta\\ X_1\cap \ldots \cap X_s=\emptyset}} \prod_{i=1}^{s} qm_i(X_i)$$

\section{Qualitative DSm Classic rule}

The qualitative DSm Classic rule (q-DSmC)\index{qualitative DSm Classic rule (q-DSmC)} for $s\geq 2$  is defined similarly to DSm Classic fusion rule (DSmC) as follows : $qm_{qDSmC}(\emptyset)=L_0$ and for all $X\in D^\Theta\setminus \{\emptyset\}$,

\begin{equation}
qm_{qDSmC}(X)=\sum_{\substack{X_1,,\ldots,X_s\in D^\Theta\\ X_1\cap\ldots \cap X_s=X}} \prod_{i=1}^{s} qm_i(X_i)
\label{qDSmC}
\end{equation}

\section{Qualitative DSm Hybrid rule}

The qualitative DSm Hybrid rule (q-DSmH)\index{qualitative DSm Hybrid rule (q-DSmH)} is defined similarly to quantitative DSm hybrid rule \cite{DSmTBook_2004a} as follows: $qm_{qDSmH}(\emptyset)=L_0$ and for all $X\in G^\Theta\setminus \{\emptyset\}$
\begin{equation}
qm_{qDSmH}(X)\triangleq
\phi(X)\cdot\Bigl[ qS_1(X) + qS_2(X) + qS_3(X)\Bigr]
 \label{qDSmH}
\end{equation}
\noindent
where all sets involved in formulas are in the canonical form and $\phi(X)$ is the {\it{characteristic non-emptiness function}} of a set $X$, i.e. $\phi(X)= L_{m+1}$ if  $X\notin \boldsymbol{\emptyset}$ and $\phi(X)= L_0$ otherwise, where $\boldsymbol{\emptyset}\triangleq\{\boldsymbol{\emptyset}_{\mathcal{M}},\emptyset\}$. $\boldsymbol{\emptyset}_{\mathcal{M}}$ is the set  of all elements of $D^\Theta$ which have been forced to be empty through the constraints of the model $\mathcal{M}$ and $\emptyset$ is the classical/universal empty set. $qS_1(X)\equiv qm_{qDSmC}(X)$, $qS_2(X)$, $qS_3(X)$ are defined by
\begin{equation}
qS_1(X)\triangleq \sum_{\substack{X_1,X_2,\ldots,X_s\in D^\Theta\\ X_1\cap X_2\cap\ldots\cap X_s=X}} \prod_{i=1}^{s} qm_i(X_i)
\end{equation}
\begin{equation}
qS_2(X)\triangleq \sum_{\substack{X_1,X_2,\ldots,X_s\in\boldsymbol{\emptyset}\\ [\mathcal{U}=X]\vee [(\mathcal{U}\in\boldsymbol{\emptyset}) \wedge (X=I_t)]}} \prod_{i=1}^{s} qm_i(X_i)
\end{equation}
\begin{equation}
qS_3(X)\triangleq\sum_{\substack{X_1,X_2,\ldots,X_k\in D^\Theta \\ 
X_1\cup X_2\cup \ldots \cup X_s=X \\ X_1\cap X_2\cap \ldots\cap X_s \in\boldsymbol{\emptyset}}}  \prod_{i=1}^{s} qm_i(X_i)
\end{equation}
\noindent
with $\mathcal{U}\triangleq u(X_1)\cup \ldots \cup u(X_s)$ where $u(X)$ is the union of all $\theta_i$ that compose $X$, $I_t \triangleq \theta_1\cup \ldots\cup \theta_n$ is the total ignorance. $qS_1(X)$ is nothing but the qDSmC rule for $s$ independent sources based on $\mathcal{M}^f(\Theta)$; $qS_2(X)$ is the qualitative mass\index{qualitative mass} of all relatively and absolutely empty sets which is transferred to the total or relative ignorances associated with non existential constraints (if any, like in some dynamic problems); $qS_3(X)$ transfers the sum of relatively empty sets directly onto the canonical disjunctive form of non-empty sets. qDSmH generalizes qDSmC works for any models (free DSm model, Shafer's model or any hybrid models) when manipulating qualitative belief assignments\index{qualitative belief assignment}.

 \section{Qualitative Average Operator}

The Qualitative Average Operator (QAO)\index{QAO (Qualitative Average Operator)} is an extension of Murphy's numerical average operator \cite{Murphy_2000}. But here we define two types of QAO's:

\begin{itemize}
\item[a)] A pessimistic (cautious) one : 
\begin{equation}
QAO_{p}(L_i,L_j)= L_{\lfloor\frac{i+j}{2}\rfloor}
\label{QAOp}
\end{equation}
\noindent
where $\lfloor x \rfloor$ means the lower integer part of $x$, i.e. the greatest integer less than or equal to $x$;
\item[a)] An optimistic one : 
\begin{equation}
QAO_{o}(L_i,L_j)= L_{\lceil\frac{i+j}{2}\rceil}
\label{QAOo}
\end{equation}
where $\lceil x \rceil$ means the upper integer part of $x$, i.e. the smallest integer greater than or equal to $x$.
\end{itemize}
\noindent
QAO can be generalized for $s\geq 2$ qualitative sources.

\section{Qualitative PCR5 rule (q-PCR5)\index{qualitative PCR5 rule (q-PCR5)}}

In classical (i.e. quantitative) DSmT\index{Dezert-Smarandache Theory (DSmT)} framework, the Proportional Conflict Redistribution rule no.~5 (PCR5) defined in \cite{Book_2006} has been proven to provide very good and coherent results for combining (quantitative) belief masses, see \cite{Smarandache_2005c,Dezert_2006b}. When dealing with qualitative beliefs and using Dempster-Shafer Theory (DST)\index{Dempster-Shafer Theory (DST)}, we unfortunately cannot normalize, since it is not possible to divide linguistic labels\index{linguistic labels} by linguistic labels\index{linguistic labels}. Previous authors have used the un-normalized Dempster's rule\index{Dempster's rule}, which actually is equivalent to the Conjunctive Rule in Shafer's model and respectively to DSm conjunctive rule in hybrid and free DSm models. Following the idea of (quantitative) PCR5 fusion rule, we can however use a rough approximation for a qualitative version of PCR5 (denoted qPCR5) as it will be presented in next example, but we did not succeed so far to get a general formula for qualitative PCR5 fusion rule (q-PCR5) because the division of labels could not be defined.

\section{A simple example}

Let's consider the following set of ordered linguistic labels\index{linguistic labels} $L=\{L_0,L_1,L_2,L_3,L_4,L_{5}\}$ (for example, $L_1$, $L_2$, $L_3$ and $L_4$ may represent the values: $L_1\triangleq \text{{\it{very poor}}}$, $L_2\triangleq \text{{\it{poor}}}$, $L_3\triangleq \text{{\it{good}}}$ and $L_4\triangleq \text{{\it{very good}}}$, where $\triangleq$ symbol means  {\it{by definition}}), then addition and multiplication tables are:

\begin{table}[h]
\centering 
\begin{tabular}{|c|cccccc|}
\hline
$+$     & $L_0$ & $L_1$ & $L_2$ & $L_3$ & $L_4$ & $L_5$\\
\hline
$L_0$ & $L_0$ & $L_1$ & $L_2$ & $L_3$ & $L_4$ & $L_5$\\
$L_1$ & $L_1$ & $L_2$ & $L_3$ & $L_4$ & $L_5$ & $L_5$\\
$L_2$ & $L_2$ & $L_3$ & $L_4$ & $L_5$ & $L_5$ & $L_5$\\
$L_3$ & $L_3$ & $L_4$ & $L_5$ & $L_5$ & $L_5$ & $L_5$\\
$L_4$ & $L_4$ & $L_5$ & $L_5$ & $L_5$ & $L_5$ & $L_5$\\
$L_5$ & $L_5$ & $L_5$ & $L_5$ & $L_5$ & $L_5$ & $L_5$\\
\hline
\end{tabular}
\caption{Addition table}
\label{CWTable3}
\end{table}
\begin{table}[h]
\centering 
\begin{tabular}{|c|cccccc|}
\hline
$\times$ & $L_0$ & $L_1$ & $L_2$ & $L_3$ & $L_4$ & $L_5$\\
\hline
$L_0$ & $L_0$ & $L_0$ & $L_0$ & $L_0$ & $L_0$ & $L_0$\\
$L_1$ & $L_0$ & $L_1$ & $L_1$ & $L_1$ & $L_1$ & $L_1$\\
$L_2$ & $L_0$ & $L_1$ & $L_2$ & $L_2$ & $L_2$ & $L_2$\\
$L_3$ & $L_0$ & $L_1$ & $L_2$ & $L_3$ & $L_3$ & $L_3$\\
$L_4$ & $L_0$ & $L_1$ & $L_2$ & $L_3$ & $L_4$ & $L_4$\\
$L_5$ & $L_0$ & $L_1$ & $L_2$ & $L_3$ & $L_4$ & $L_5$\\
\hline
\end{tabular}
\caption{Multiplication table}
\label{CWTable4}
\end{table}

\newpage
\noindent The tables for $QAO_{p}$ and $QAO_{o}$ operators are:

\begin{table}[h]
\centering 
\begin{tabular}{|c|cccccc|}
\hline
$QAO_{p}$ & $L_0$ & $L_1$ & $L_2$ & $L_3$ & $L_4$ & $L_5$\\
\hline
$L_0$ & $L_0$ & $L_0$ & $L_1$ & $L_1$ & $L_2$ & $L_2$\\
$L_1$ & $L_0$ & $L_1$ & $L_1$ & $L_2$ & $L_2$ & $L_3$\\
$L_2$ & $L_1$ & $L_1$ & $L_2$ & $L_2$ & $L_3$ & $L_3$\\
$L_3$ & $L_1$ & $L_2$ & $L_2$ & $L_3$ & $L_3$ & $L_4$\\
$L_4$ & $L_2$ & $L_2$ & $L_3$ & $L_3$ & $L_4$ & $L_4$\\
$L_5$ & $L_2$ & $L_3$ & $L_3$ & $L_4$ & $L_4$ & $L_5$\\
\hline
\end{tabular}
\caption{Table for $QAO_{p}$}
\label{TableQAOp}
\end{table}
\begin{table}[!h]
\centering 
\begin{tabular}{|c|cccccc|}
\hline
$QAO_{o}$ & $L_0$ & $L_1$ & $L_2$ & $L_3$ & $L_4$ & $L_5$\\
\hline
$L_0$ & $L_0$ & $L_1$ & $L_1$ & $L_2$ & $L_2$ & $L_3$\\
$L_1$ & $L_1$ & $L_1$ & $L_2$ & $L_2$ & $L_3$ & $L_3$\\
$L_2$ & $L_1$ & $L_2$ & $L_2$ & $L_3$ & $L_3$ & $L_4$\\
$L_3$ & $L_2$ & $L_2$ & $L_3$ & $L_3$ & $L_4$ & $L_4$\\
$L_4$ & $L_2$ & $L_3$ & $L_3$ & $L_4$ & $L_4$ & $L_5$\\
$L_5$ & $L_3$ & $L_3$ & $L_4$ & $L_4$ & $L_5$ & $L_5$\\
\hline
\end{tabular}
\caption{Table for $QAO_{o}$}
\label{TableQAOo}
\end{table}

Let's consider now a simple two-source case with a 2D frame $\Theta=\{\theta_1,\theta_2\}$, Shafer's model for $\Theta$, and qba's expressed as follows:
$$qm_1(\theta_1)=L_1, \quad qm_1(\theta_2)=L_3, \quad qm_1(\theta_1\cup\theta_2)=L_1$$
$$qm_2(\theta_1)=L_2, \quad qm_2(\theta_2)=L_1, \quad qm_2(\theta_1\cup\theta_2)=L_2$$

\subsection{Qualitative Fusion of qba's}

\begin{itemize}
\item {\bf{Fusion with (qCR)}}: According to qCR combination rule \eqref{qCR}, one gets the result in Table \ref{CWTable5}, since
\begin{align*}
qm_{qCR}(\theta_1) &=qm_1(\theta_1)qm_2(\theta_1) + qm_1(\theta_1)qm_2(\theta_1\cup\theta_2)  + qm_2(\theta_1)qm_1(\theta_1\cup\theta_2)\\
&= (L_1\times L_2)+(L_1\times L_2)+(L_2\times L_1)\\
&= L_1+L_1+L_1 = L_{1+1+1}=L_3
\end{align*}
\begin{align*}
qm_{qCR}(\theta_2) &=qm_1(\theta_2)qm_2(\theta_2) + qm_1(\theta_2)qm_2(\theta_1\cup\theta_2)  + qm_2(\theta_2)qm_1(\theta_1\cup\theta_2)\\
&= (L_3\times L_1)+(L_3\times L_2)+(L_1\times L_1)\\
&= L_1+L_2+L_1 = L_{1+2+1}=L_4
\end{align*}
\begin{align*}
qm_{qCR}(\theta_1\cup\theta_2) & = qm_1(\theta_1\cup\theta_2)qm_2(\theta_1\cup\theta_2)= L_1\times L_2 = L_1
\end{align*}
\begin{align*}
qm_{qCR}(\emptyset) & \triangleq K_{12} =  qm_1(\theta_1)qm_2(\theta_2)+ qm_1(\theta_2)qm_2(\theta_1)\\
&= (L_1\times L_1) +  (L_2\times L_3) = L_1+L_2= L_3
\end{align*}
\end{itemize}
In summary, one gets
\begin{table}[!h]
\centering
\begin{tabular}{|l|ccccc|}
\hline
 & $\theta_1$ & $\theta_2$ & $\theta_1\cup\theta_2$ & $\emptyset$ & $\theta_1\cap\theta_2$\\
\hline
$qm_1(.)$ & $L_1$ & $L_3$ & $L_1$ &  &  \\
$qm_2(.)$ & $L_2$ & $L_1$ & $L_2$ &  &  \\
\hline
$qm_{qCR}(.)$ & $L_3$ & $L_{4}$ & $L_1$ &  $L_3$ &  $L_0$\\
\hline
\end{tabular}
\caption{Fusion with qCR}
\label{CWTable5}
\end{table}
\newpage

\begin{itemize}
\item {\bf{Fusion with (qDSmC)}}: If we accepts the free-DSm model instead Shafer's model, according to qDSmC combination rule \eqref{qDSmC}, one gets the result in Table \ref{TableDSmC},
\end{itemize}

\begin{table}[h]
\centering
\begin{tabular}{|l|ccccc|}
\hline
 & $\theta_1$ & $\theta_2$ & $\theta_1\cup\theta_2$ & $\emptyset$ & $\theta_1\cap\theta_2$\\
\hline
$qm_1(.)$ & $L_1$ & $L_3$ & $L_1$ &  &  \\
$qm_2(.)$ & $L_2$ & $L_1$ & $L_2$ &  &  \\
\hline
$qm_{qDSmC}(.)$ & $L_3$ & $L_4$ & $L_1$ &  $L_0$ &  $L_3$\\
\hline
\end{tabular}
\caption{Fusion with qDSmC}
\label{TableDSmC}
\end{table}

\begin{itemize}
\item {\bf{Fusion with (qDSmH)}}: Working with Shafer's model for $\Theta$, according to qDSmH combination rule \eqref{qDSmH}, one gets the result in Table \ref{TableDSmH}.
\end{itemize}

\begin{table}[h]
\centering
\begin{tabular}{|l|ccccc|}
\hline
 & $\theta_1$ & $\theta_2$ & $\theta_1\cup\theta_2$ & $\emptyset$ & $\theta_1\cap\theta_2$\\
\hline
$qm_1(.)$ & $L_1$ & $L_3$ & $L_1$ &  &  \\
$qm_2(.)$ & $L_2$ & $L_1$ & $L_2$ &  &  \\
\hline
$qm_{qDSmH}(.)$ & $L_3$ & $L_4$ & $L_4$ &  $L_0$ &  $L_0$\\
\hline
\end{tabular}
\caption{Fusion with qDSmC}
\label{TableDSmH}
\end{table}

since $qm_{qDSmH}(\theta_1\cup\theta_2)=L_1+L_3=L_4$.

\begin{itemize}
\item {\bf{Fusion with QAO}}: Working with Shafer's model for $\Theta$, according to QAO combination rules \eqref{QAOp} and \eqref{QAOo}, one gets the result in Table \ref{QAOFusionResult}.
\end{itemize}

\begin{table}[!h]
\centering 
\begin{tabular}{|l|ccc|}
\hline
 & $\theta_1$ & $\theta_2$ & $\theta_1\cup\theta_2$ \\
\hline
$qm_1(.)$ & $L_1$ & $L_3$ & $L_1$ \\
$qm_2(.)$ & $L_2$ & $L_1$ & $L_2$ \\
\hline
$qm_{QAO_p}(.)$ & $L_1$ & $L_2$ & $L_1$\\
$qm_{QAO_o}(.)$ & $L_2$ & $L_2$ & $L_2$\\
\hline
\end{tabular}
\caption{Fusion of qba's with QAO's}
\label{QAOFusionResult}
\end{table}

\begin{itemize}
\item {\bf{Fusion with (qPCR5)}}:
Following PCR5 method, we propose to transfer the qualitative partial masses
\begin{enumerate}
\item[a)] $qm_1(\theta_1)qm_2(\theta_2)=L_1\times L_1=L_1$ to $\theta_1$ and $\theta_2$ in equal parts (i.e. proportionally to $L_1$ and $L_1$ respectively, but $L_1=L_1$); hence $\frac{1}{2}L_1$ should go to each of them.
\item[b)] $qm_2(\theta_1)qm_1(\theta_2)=L_2\times L_3=L_2$ to $\theta_1$ and $\theta_2$ proportionally to $L_2$ and $L_3$ respectively; but since we are not able to do an exact proportionalization of labels, we approximate through transferring  $\frac{1}{3}L_2$ to $\theta_1$ and $\frac{2}{3}L_2$ to $\theta_2$.
\end{enumerate}

The transfer $(1/3) L_2$ to $\theta_1$ and $(2/3) L_2$ to $\theta_2$ is not arbitrary, but it is an approximation since the
transfer was done proportionally to $L_2$ and $L_3$, and $L_2$ is smaller than $L_3$; we mention that it is not possible to do an exact transferring. Nobody in the literature has done so far normalization of labels, and we tried to do a quasi-normalization\index{quasi-normalization} [i.e. an approximation].\\

Summing a) and b) we get: $\frac{1}{2}L_1 + \frac{1}{3}L_2\approx L_1$, which represents the partial conflicting qualitative mass\index{qualitative mass} transferred to $\theta_1$, and $\frac{1}{2}L_1 +  \frac{2}{3}L_2\approx L_2$, which represents the partial conflicting qualitative mass\index{qualitative mass} transferred to $\theta_2$. Here we have mixed qualitative and quantitative information.

Hence we will finally get:
\begin{table}[h]
\centering
\begin{tabular}{|l|ccccc|}
\hline
 & $\theta_1$ & $\theta_2$ & $\theta_1\cup\theta_2$ & $\emptyset$ & $\theta_1\cap\theta_2$\\
\hline
$qm_1(.)$ & $L_1$ & $L_3$ & $L_1$ &  &  \\
$qm_2(.)$ & $L_2$ & $L_1$ & $L_2$ &  &  \\
\hline
$qm_{qPCR5}(.)$ & $L_4$ & $L_5$ & $L_1$ &  $L_0$ &  $L_0$\\
\hline
\end{tabular}
\caption{Fusion with qPCR5}
\label{TableqPCR5}
\end{table}

\noindent
For the reason that we can not do a normalization (neither previous authors on qualitative fusion rules\index{qualitative fusion rules} did), we propose for the first time the possibility of {\it{quasi-normalization\index{quasi-normalization}}} (which is
an approximation of the normalization), i.e. instead of dividing each qualitative mass\index{qualitative mass} by a coefficient of normalization, we {\it{subtract}} from each qualitative mass\index{qualitative mass} a qualitative coefficient (label) of quasi-normalization\index{quasi-normalization} in order to adjust the sum of masses.\\

Subtraction on $L$ is defined in a similar way to the addition:
\begin{equation}
L_i - L_j=
\begin{cases}
L_{i-j}, &\quad\text{if} \ i\geq j;\\
L_0, &\quad\text{if} \ i< j;
\end{cases}
\label{qsub}
\end{equation}
\noindent
$L$ is closed under subtraction as well.\\

The subtraction can be used for quasi-normalization\index{quasi-normalization} only, i. e. moving the final label result 1-2 steps/labels up or down.  It is not used together with addition or multiplication.\\

The increment in the sum of fusioned qualitative masses\index{qualitative mass} is due to the fact that multiplication on $L$ is approximated by a larger number, because multiplying any two numbers $a$, $b$ in the interval $[0,1]$, the product is less than each of them, or we have approximated the product $a\times b = \min\{a,b\}$.

\noindent
Using the quasi-normalization\index{quasi-normalization} (subtracting $L_1$), one gets with qDSmH and qPCR5, the following {\it{quasi-normalized}} masses (we use $\star$ symbol to specify the quasi-normalization\index{quasi-normalization}):

\begin{table}[h]
\centering
\begin{tabular}{|l|ccccc|}
\hline
 & $\theta_1$ & $\theta_2$ & $\theta_1\cup\theta_2$ & $\emptyset$ & $\theta_1\cap\theta_2$\\
\hline
$qm_1(.)$ & $L_1$ & $L_3$ & $L_1$ &  &  \\
$qm_2(.)$ & $L_2$ & $L_1$ & $L_2$ &  &  \\
\hline
$qm_{qDSmH}^\star (.)$ & $L_2$ & $L_3$ & $L_3$ &  $L_0$ &  $L_0$\\
$qm_{qPCR5}^\star (.)$ & $L_3$ & $L_4$ & $L_0$ &  $L_0$ &  $L_0$\\
\hline
\end{tabular}
\caption{Fusion with quasi-normalization\index{quasi-normalization}}
\label{Tableqn}
\end{table}
\end{itemize}

\subsection{Fusion with a crisp mapping}

If we consider the labels as equidistant, then we can divide the whole interval [0,1] into five equal parts, hence mapping the linguistic labels\index{linguistic labels} $L_i$ onto crisp numbers as follows:
$$L_0 \mapsto 0, L_1 \mapsto 0.2, L_2 \mapsto 0.4, L_3 \mapsto 0.6, L_4 \mapsto 0.8, L_5 \mapsto 1$$
Then the qba's $qm_1(.)$ and $qm_2(.)$ reduce to classical (precise) quantitative belief masses $m_1(.)$ and $m_2(.)$. In our simple example, one gets
$$m_1(\theta_1)=0.2 \qquad m_1(\theta_2)=0.6 \qquad m_1(\theta_1\cup\theta_2)=0.2$$
$$m_2(\theta_1)=0.4 \qquad m_2(\theta_2)=0.2 \qquad m_2(\theta_1\cup \theta_2)=0.4$$

We can apply any classical (quantitative) fusion rules. For example, with quantitative Conjunctive Rule, Dempster-Shafer (DS), DSmC, DSmH, PCR5 and Murphy's (Average Operator - AO) rules, one gets the results in Tables \ref{TableCrisp1} and \ref{TableCrisp}.

\begin{table}[h]
\centering 
\begin{tabular}{|l|ccc|}
\hline
 & $\theta_1$ & $\theta_2$ & $\theta_1\cup\theta_2$  \\
\hline
$m_1(.)$ & $0.2$ & $0.6$ & $0.2$ \\
$m_2(.)$ & $0.4$ & $0.2$ & $0.4$ \\
\hline
$m_{CR}(.)$       & $0.24$ & $0.40$ & $0.08$ \\
$m_{DSmC}(.)$ & $0.24$ & $0.40$ & $0.08$ \\
\hline
$m_{DS}(.)$ & $\approx 0.333$ & $\approx 0.555$ & $\approx 0.112$ \\
$m_{DSmH}(.)$ & $0.24$ & $0.40$ & $0.36$ \\
$m_{PCR5}(.)$ & $0.356$ & $0.564$ & $0.080$ \\
$m_{AO}(.)$ & $0.3$ & $0.4$ & $0.3$ \\
\hline
\end{tabular}
\caption{Fusion through a crisp mapping}
\label{TableCrisp1}
\end{table}

\begin{table}[h]
\centering 
\begin{tabular}{|l|cc|}
\hline
 &  $\emptyset$ & $\theta_1\cap\theta_2$\\
\hline
$m_1(.)$ &  &  \\
$m_2(.)$ &  &  \\
\hline
$m_{CR}(.)$       & $0.28$ &  $0$\\
$m_{DSmC}(.)$ & $0$ &  $0.28$\\
\hline
$m_{DS}(.)$ &  $0$ &  $0$\\
$m_{DSmH}(.)$ &  $0$ &  $0$\\
$m_{PCR5}(.)$ &  $0$ &  $0$\\
$m_{AO}(.)$ &  $0$ &  $0$\\
\hline
\end{tabular}
\caption{Fusion through a crisp mapping (cont'd)}
\label{TableCrisp}
\end{table}

\noindent
{\bf{Important remark}}: The mapping of linguistic labels\index{linguistic labels} $L_i$ into crisp numbers $x_i\in [0,1]$ is a very difficult problem in general since the crisp mapping must generate from qualitative belief masses $qm_i(.)$, $i=1,\ldots,s$, a set of complete normalized precise quantitative belief masses $m_i(.)$, $i=1,\ldots,s$ (i.e. a set of crisp numbers in $[0,1]$ such $\sum_{X\in G^\Theta} m_i(X)=1$, $\forall i=1,\ldots,s$). According to \cite{Wong_1991,Wong_1992}, such direct crisp mapping function can be found/built only if the qba's satisfy a given set of constraints. Generally a crisp mapping function and qba's generate for at least one of sources to combine either a paraconsistent (quantitative) belief assignments (if the sum of quantitative masses is greater than one) or an incomplete belief assignment (if the sum of masses is less than one). An issue would be in such cases to make a (quantitative) normalization of  all paraconsistent and  incomplete belief assignments drawn from crisp mapping and qba's before combining them with a quantitative fusion rule. The normalization of paraconsistent and incomplete bba's reflects somehow the difference in the interpretations of labels used by the sources (i.e. each source carries its own (mental/internal) representation of the linguistic label he/she uses when committing qualitative beliefs on any given proposition). It is possible to approximate the labels by crisp numbers of by subunitary subsets (in imprecise information), but the accuracy is arguable.

\subsection{Fusion with an interval mapping}

An other issue to avoid the direct manipulation of qualitative belief masses, is to try to assign intervals assign intervals or more general subunitary subsets to linguistic labels\index{linguistic labels} in order to model the vagueness of labels into numbers. We call this process, the {\it{interval mapping}} of qba's. This approach is less precise than the crisp mapping approach but is a quite good compromise between qualitative belief fusion and (precise) quantitative belief fusion.\\

\noindent
In our simple example, we can easily check that the following interval mapping
$$L_0\mapsto [0,0.1), L_1 \mapsto [0.1,0.3), L_2 \mapsto [0.3,0.5),L_3 \mapsto [0.5,0.7),L_4 \mapsto [0.7,0.9), L_5 \mapsto [0.9,1]$$

\noindent
allows us to build two set of admissible\footnote{Admissibility condition means that we can pick up at least one number in each interval of an imprecise belief mass in such a way that the sum of these numbers is one (see \cite{DSmTBook_2004a} for details and examples). For example, $m_1^I(.)$ is admissible since there exist $0.22\in [0.1, 0.3)$, $0.55 \in [0.5, 0.7)$, and $0.23\in [0.1, 0.3)$ such that $0.22+0.55+0.23=1$.} imprecise (quantitative) belief masses:
$$m_1^I(\theta_1)=[0.1,0.3) \qquad m_2^I(\theta_1)=[0.3,0.5)$$
$$m_1^I(\theta_2)=[0.5,0.7) \qquad m_2^I(\theta_2)=[0.1,0.3)$$
$$m_1^I(\theta_1\cup\theta_2)=[0.1,0.3)\qquad m_2^I(\theta_1\cup \theta_2)=[0.3,0.5)$$

These two admissible imprecise belief assignments can then be combined with (imprecise) combination rules proposed in \cite{DSmTBook_2004a} and based on the following operators for interval calculus:  If $\mathcal{X}_{1}$,$\mathcal{X}_{2}$,\ldots, $\mathcal{X}_{n}$ are real sets, then their
sum is:
 \begin{equation*}
 \underset{k=1,\ldots,n}{\boxed{\sum}} \mathcal{X}_{k} = \{ x \mid x = \sum_{k=1,\ldots,n} x_{k},  x_{1} \in 
 \mathcal{X}_{1},\ldots, x_{n} \in  \mathcal{X}_{n}\}
  \end{equation*}
  \noindent
while their product is:
\begin{equation*}
 \underset{k=1,\ldots,n}{\boxed{\prod}} \mathcal{X}_{k} = \{ x \mid x = \prod_{k=1,\ldots,n} x_{k},  x_{1} \in 
 \mathcal{X}_{1},\ldots, x_{n} \in  \mathcal{X}_{n}\}
  \end{equation*}

The results obtained with an interval mapping for the different (quantitative) rules of combination are summarized in Tables \ref{ImpFusionResult1} and \ref{ImpFusionResult}.

\begin{table}[h]
\centering 
\begin{tabular}{|l|ccc|}
\hline
 & $\theta_1$ & $\theta_2$ & $\theta_1\cup\theta_2$ \\
\hline
$m_1^I(.)$ & $[0.1,0.3)$ & $[0.5,0.7)$ & $[0.1,0.3)$  \\
$m_2^I(.)$ & $[0.3,0.5)$ & $[0.1,0.3)$ & $[0.3,0.5)$\\
\hline
$m_{CR}^I(.)$       & $[0.09,0.45)$ & $[0.21,0.65)$ & $[0.03,0.15)$ \\
$m_{DSmC}^I(.)$ & $[0.09,0.45)$ & $[0.21,0.65)$ & $[0.03,0.15)$ \\
\hline
$m_{DSmH}^I(.)$ & $[0.09,0.45)$ & $[0.21,0.65)$ & $[0.19,0.59)$ \\
$m_{PCR5}^I(.)$ & $[0.15125,0.640833)$ & $[0.30875,0.899167)$ & $[0.03, 0.15)$\\
$m_{AO}^I(.)$ & $[0.2,0.4)$ & $[0.3,0.5)$ & $[0.2,0.4)$ \\
\hline
\end{tabular}
\caption{Fusion Results with interval mapping}
\label{ImpFusionResult1}
\end{table}
\begin{table}[h]
\centering 
\begin{tabular}{|l|cc|}
\hline
 &  $\emptyset$ &  $\theta_1\cap\theta_2$\\
\hline
$m_1^I(.)$ &  &  \\
$m_2^I(.)$ &  & \\
\hline
$m_{CR}^I(.)$       & $[0.16,0.44)$ &  $0$\\
$m_{DSmC}^I(.)$ & $0$ &  $[0.16,0.44)$\\
\hline
$m_{DSmH}^I(.)$ &  $0$ &  $0$\\
$m_{PCR5}^I(.)$  &  $0$ &  $0$\\
$m_{AO}^I(.)$ &  $0$ &  $0$\\
\hline
\end{tabular}
\caption{Fusion Results with interval mapping (cont'd)}
\label{ImpFusionResult}
\end{table}

\section{Conclusion}

We have extended in this paper the use of DSmT\index{Dezert-Smarandache Theory (DSmT)} from quantitative to qualitative belief assignments\index{qualitative belief assignment}. In order to apply the fusion rules to qualitative information, we defined the $+$, $\times$, and even $-$ operators working on the set of linguistic labels\index{linguistic labels}. Tables of qualitative calculations are presented and examples using the corresponding qualitative-type Conjunctive, DSm Classic, DSm Hybrid, PCR5 rules, and qualitative-type Average Operator. We also mixed the qualitative and quantitative information in an attempt to refine the set of linguistic labels\index{linguistic labels} for a better accuracy.  Since a normalization is not possible because the division of labels does not work, we introduced a quasi-normalization\index{quasi-normalization} (i.e. approximation of the normalization). Then mappings were designed from qualitative to (crisp or interval) quantitative belief assignments.

\end{document}